# Effective XML Representation for Spoken Language in Organisations

Rodney J. Clarke, Philip C. Windridge and Dali Dong

Faculty of Computing, Engineering and Technology, Staffordshire University
Beaconside, Stafford, United Kingdom ST180DG
{ R.J.Clarke | P.C.Windridge | D.Dong }@staffs .ac.uk

**Abstract**

Spoken Language can be used to provide insights into organisational processes, unfortunately transcription and coding stages are very time consuming and expensive. The concept of partial transcription and coding is proposed in which spoken language is indexed prior to any subsequent processing. The functional linguistic theory of texture is used to describe the effects of partial transcription on observational records. The standard used to encode transcript context and metadata is called CHAT, but a previous XML schema developed to implement it contains design assumptions that make it difficult to support partial transcription for example. This paper describes a more effective XML schema that overcomes many of these problems and is intended for use in applications that support the rapid development of spoken language deliverables.

## 1.     Significance of Spoken Language Resources

Spoken Language is a central but nonetheless often overlooked organisational resource. Its complexity signals its enormous potential for a variety of organisational applications including but not limited to the analysis of decision-making processes; negotiations occurring during the introduction of new work practices into work places; training and deployment of methods; and systems analysis, design, development, installation, operation and decommissioning. The complexity of spoken language can be studied using a variety of research approaches including various kinds of qualitative analysis, contextual analysis, ethnography, semiotics, and linguistics. As spoken language is used to represent different kinds of meanings than written language, this will necessitate special technologies for its potential to be fully appreciated and used.

A major operational difficulty that affects the acceptance and uptake of approaches that utilise spoken language resources to study organisations and their associated technologies is the performance bottleneck associated with the transcription and coding processes. In part this is a consequence of some basic assumptions about what constitutes adequate spoken language data from a research perspective. The belief that transcripts can only be of use when they completely cover the entire observational record and are exhaustive in terms of coding is referred to here as a *monolithic* view of transcription and its deliverables. As transcripts can be re-analysed or reused for different purposes, the notion that coding in particular can ever be complete is questionable. Furthermore, due consideration must be given to issues of security, privacy, confidentiality and intellectual property related to spoken language in organisational settings. It must also be acknowledged that organisations constitute 'unsafe environments' for participants (Cameron et al 1992) involving issues of access, control, power and representation. Therefore, the assumptions that inform a monolithic view of transcription and coding deliverables may need to be revised. In the following section we propose moving from a monolithic to a 'partial' view of transcription and coding, and suggest theory and methods that can assist us in understanding the consequences of doing so.

## 2.    Concept of Partial Transcription

One obvious strategy for dealing with the bottlenecks and problems associated with transcription and coding processes in organisational settings is to omit chunks of the observational record based on the occurrence of an explicit *indexing phase* prior to transcription and coding itself. In contrast to a monolithic approach to transcription, previously described, the production of intentionally incomplete transcription and coding deliverables is referred to here as *partial transcription*. The advantages of partial transcription include amongst other things the ability to encourage empowerment research (see Cameron et al 1992) by enabling the participants themselves to determine what gets recorded. This facilitates trust and improves the research relationship between analysts and members of organisations. An obvious difficulty with partial transcription is that, depending on the kind of analysis being undertaken, omitting sections of a transcript will disrupt a number of spoken language resources- some of these may be crucial to the analysis being conducted. Fortunately, functional linguistic theories exist which can give considerable insight into which specific language resources will be affected. We use Systemic Functional Linguistics (SFL) a semiotic model of language (Halliday 1985) because it has a concept referred to as *texture* that encompasses and defines all the text-forming resources that may be used in a transcript or any other text. For example, texture has also been applied to hypertext development and modeling (Clarke 1997). Any texts including all transcripts must possess texture in order to function as a semantic unit, as well as being relevant or appropriate to a given social setting or occasion. Whether knowingly or not, speakers and writers use their experience of texture resources when constructing texts, while listeners and readers use their experience of these resources when interpreting texts. Texts are generally read from start to finish and so many of these resources flow through a text in chains. This is an attribute of language referred to as *sequential implicativeness* (Schegloff and Sachs 1973). For example a text might start with the sentence "Rod is in the Red Theatre" and if the next sentence was "He is giving a seminar" we might reasonably conclude that the 'He' is Rod. This is an effect of sequential implicativeness in the so-called Reference System (see below).

There are several models of texture within SFL. The texture model we use (Martins' 1992, 381 adaptation of Halliday and Hasan's 1976 model) recognises three major groups of text-forming resources- Intrasentential Resources, Intersentential Resources, and

Coherence. Within each major group, there are a number of sub-categories of text-forming resources each having an associated analysis method and some also have graphical methods:

- intra-sentential resources (Martin 1992, 381) or *structural resources* (Halliday 1985)- involve systems of THEME and INFORMATION and spoken language specific systems involved in Conversation Structure. All texts consist of sets of clauses each of which can be divided into a *theme* and a *rheme*. Listeners or readers rely upon *thematic progression*, the specific pattern of themes, to predict how the text should unfold. Texts must also provide and 'manage' information. Listeners or readers come to rely upon patterns of information units to build and accumulate *new* meanings from those that have already been *given*. Conversation Structure involves *speech functions* the characteristic set of moves enacted by participants involving initiations (offers, commands, statements, questions) or responses, as well as sequences of speech functions that form jointly negotiated patterns called e*xchange structure*.

- intersentential text-forming resources of *Cohesion*- describe how clauses within any text are interrelated giving the appearance of a unity thereby assisting listeners and readers in understanding the meanings being negotiated. There are a number of types of cohesion, including *lexical cohesion* which describes how lexical items (words) and sequences of events are used to consistently relate a text to a topic, *reference* which describes how participants are introduced and subsequently managed, *ellipsis* which establishes reference relationships through the omission of otherwise repetitive lexical items, *substitution* which employs alternate lexis for original lexical items, and *conjunction* which refers to the logical relations between parts of a text.

- text forming resources of *Coherence*- which describes how clauses in texts relate to the contexts in which they occur. All texts must be relevant to the immediate situational context, referred to as *situational coherence*, while also conforming to an appropriate genre, referred to as *generic coherence*.

It is relatively easy to understand in principle what happens when we partially transcribe. Effectively we run the risk of disrupting sequential implicativeness of many of these text-forming resources. Partial transcripts may loose coherence, and will most certainly have disrupted thematic and informational intra-sentential resources. Perhaps the group of text forming resources most disrupted will be cohesion as omitting clauses make it more difficult for readers to understand the transcript as a unity. We could easily produce an unintelligible partial transcript if we removed too much of it. While texture theory can tell us which language resources will be affected when we adopt partial transcription, it can only provide part of the picture. The theory of texture cannot tell us how significant the disruption will be for the type of research methodology being undertaken. NLP analyses may find partial transcription useful because redundancy within and interdependency between text-forming resources can offset the fact that the transcript is not complete. We might expect qualitative analyses to be adversely affected by partial transcription although this may be almost completely offset by carefully designing the indexing phase. The indexing phase may function to provide Code Tables for those qualitative methodologies that use descriptive, interpretative or pattern codes based on relatively pre-established analytical categories (see Miles and Huberman 1994, 57-72). However, grounded theory and ethnographic methodologies are more likely to be adversely effected by adopting partial transcription. Of course, it is impossible here to consider all the ways in which a text may have its texture forming resources affected by partial transcription, but knowledge of these resources can help us greatly. Having established partial transcription as a potentially useful approach to dealing with the bottlenecks associated with transcription and coding processes in organisation it became a mandatory requirement in our studies. We now turn our attention to ways in which we represent the transcript content and metadata.

## 3.      Representing Talk: CHAT and the TalkBank Schema

Even in the research literature, transcription is often ad hoc and idiosyncratic; formal standards are not necessarily well known. One of the best-defined transcription standards is CHAT- *C*odes for the *H*uman *A*nalysis of *T*ranscripts developed by Brain MacWhinney and Jane Walter at the CHILDES- *Chi*ld *L*anguage *D*ata *E*xchange Research Centre, Department of Psychology, Carnegie Mellon University (CHILDES 2003). CHAT is a scalable, elaborate and expressive standard that supports transcription and coding even under the most adverse of conditions (participants with speech impediments, unclear or noisy recordings, breaks in the observational record). The standard is extensible, providing a consistent way of adding new headers if necessary (MacWhinney 2003). As illustrated in Figure 1, CHAT transcripts have a common basic structure. A block of so-called *Constant Headers* at the *top of the transcript* starting with an @Begin provides persistent information, which is applicable throughout the transcript. Some headers can occur more than once in a transcript signalling for example, changes in situation, space and time, and are referred to as *Changeable Headers*. The *body of the transcript* consists of speaker utterances called *Mainline*s, signalled with an asterix and a three-letter participant code. Each mainline may be followed by zero or more *Dependent Tiers*, used for coding information about the utterances. These start with a percent sign and three-letter code that indicates the type of coding information provided. The single command @End is used to mark the *end of the transcript*.

One of the reasons that CHAT is of interest for researchers of spoken language in organisational settings is that it was developed with subsequent computer processing in mind. A suite of programs called CLAN can be used to parse CHAT compliant transcripts. Development work has proceeded in several directions under the aegis of a National Science Foundation funded joint project between Carnegie Mellon and Pennsylvania Universities called TalkBank. The first direction involves expanding the range of media used in the study of communication. The second direction involves leveraging the advantages afforded by XML and related technologies. Within TalkBank, the spoken language resources are represented using CHAT. The design of the current TalkBank (2003a) schema appears to be based on creating an XML version of the CHAT standard and for the most part appears to reproduce the structure of a CHAT file itself. A design assumption that informs the TalkBank schema is that transcripts are monolithic entities, as previously defined, and this reflects the kind of applications that CHAT was developed to address. As described in section 2, transcription in organisational settings and for organisational purposes necessitates a different design approach, which we believe makes the adoption of the current TalkBank CHAT schema problematic for the following reasons:

**Figure 1**:  Structure of a simple CHAT Transcript. The special symbols in the mainlines indicate group structure. The excerpt is from 'SL6', SemLab Corpora (after Clarke et al 2003).

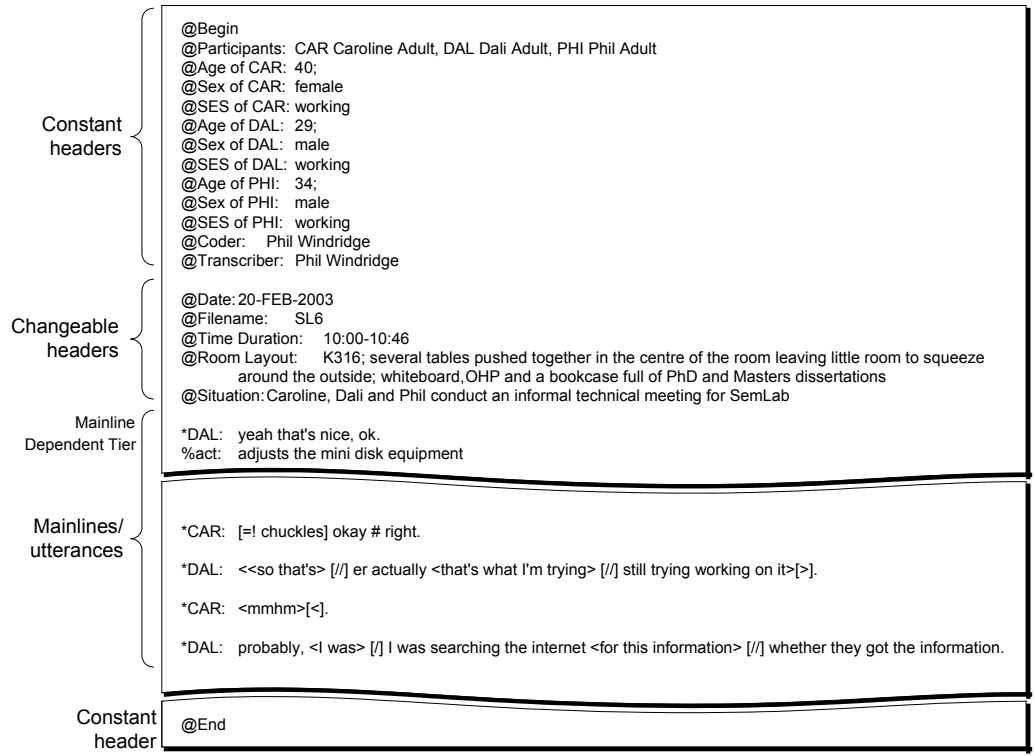

- *Monolithic View of Transcripts*: Above the level of the utterance, TalkBank views transcripts as monolithic and non-hierarchical entities. However in organisational settings, transcripts may more usefully be considered otherwise and this should be reflected in the XML used to represent them. An appropriate XML schema for these purposes would, for example, place episodes at a higher level than utterances in the XML hierarchy.

- *Overloading of the Schema*: A consequence of modeling the design of the XML schema on the CHAT file structure itself is that the resulting single file becomes huge. If all analyses must be contained within the one file then any schema will become overloaded.

- *Over-specification of CHAT semantics*: The TalkBank XML schema over-specifies certain aspects of the CHAT syntax. For example with reference to the *barb01.xml* transcript (TalkBank 2003b), the use of an explicit pause element complicates the schema by replacing one character with an entire line, while not contributing anything to the CHAT semantics.

- *Coupling between Schema and Standard*: The over-specification of the CHAT semantics, described above, induces a very tight coupling between the CHAT standard and its current XML representation. This means that any changes to the standard also require reworking of the XML schema itself. This may create a version control problem, likely to require modification of the existing XML schema, XML documents using this schema and associated XSLTs.

- *Poor Group Support*: There are several kinds of scoped symbols used in CHAT to show ranges within utterances (MacWhinney 2003). Of particular interest in our applications are paralinguistic scoping, the provision of explanations or alternative realisations for what was spoken, retracing by or overlapping between speakers, or to signal the existence of errors. Limitations in the TalkBank (2003a) schema preclude the representation of overlapping group structures, which are extremely important for NLP applications. Figure 2 shows an XML representation of an excerpt from *SL6*, part of the SemLab Corpora (Clarke et al 2003). The group structure in the excerpt can be interpreted in two entirely different ways. The region of ambiguity can only be resolved with reference to the original source material.

**Figure 2**:  XML Representation of the excerpt from 'SL6', SemLab Corpora (after Clarke et al 2003). Two entirely different interpretations of a group structure are shown (the right hand side shows an embedded group). The ambiguity between these interpretations affects the text between the dashed lines.

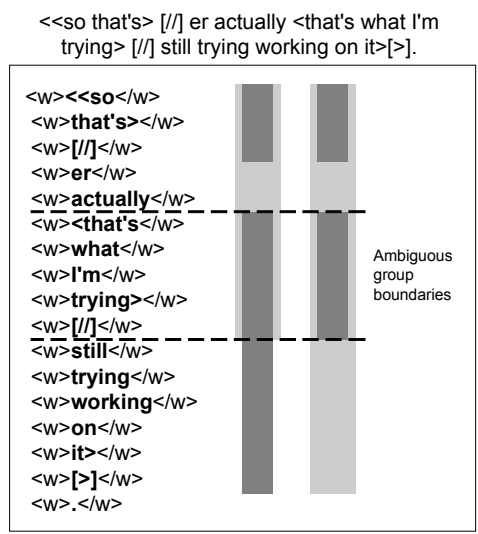

- *Lack of Version Control*: As with conventional CHAT, the TalkBank schema does not enable the content of a transcript to be version controlled. In effect the only version of a transcript is the current one. A lack of version control is a significant problem for organisational applications, which assume that spoken language resources are not monolithic.

In order to support transcription and coding in organisational settings, the current TalkBank schema was abandoned and we developed an XML schema called the Spoken Language Architecture. In order to use this new schema we developed tools which enables us to prepare, index, transcribe and code transcripts. The following section describes our XML representation for the Spoken Language Architecture.

## 4.    XML Representation for the Spoken Language Architecture

In order to develop the Spoken Language Architecture for use in organisational settings, we abandoned the design option of using a single XML file to represent a transcript, in favour of a multi-file XML schema. The resulting Spoken Language Architecture consists of a Descriptor file comprising most of the coding and indexing information that may be supplied by other applications, and a Root file that consists of utterances and the codes that are immediately related to them, see Figure 3.

**Figure 3**:   Relating the SLA Descriptor file to word group ranges in the Root file.

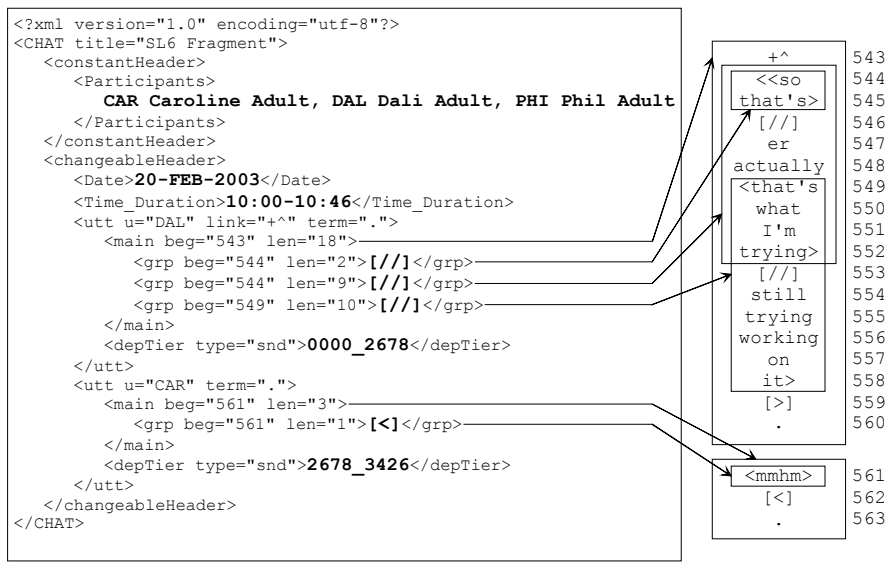

The SLA schema has the following attributes:

- *Incremental Production of Transcripts*: The concept of partial transcription was described in section 2. The multi-file XML schema is the basis of the Spoken Language Architecture and supports the *incremental development* of transcripts- akin to developing 'transcripts in pieces'.

- *Extensibility and Processing*: The XML schemas used in the Spoken Language Architecture are designed to be as open as possible- as new applications emerge for spoken language resources, the XML schema can be expanded to accommodate them by associating purpose specific Descriptor files with the SLA Root file.

- *Abstraction of CHAT Semantics*: The Spoken Language Architecture abstracts the CHAT semantics in such a way as to reduce the need to change the design of the XML when the syntax of the CHAT standard changes. An example of this is information related to participants in the transcripts. Alterations to CHAT standard syntax have resulted in the addition of an obligatory @ID header to present information in a different format. The abstraction of this information in the SLA architecture would allow reformatting to whatever syntax was applicable in a given version.

- *Partially or Completely Overlapping Regions*: Correctly identifying groups adds semantic information to the transcript. However group boundaries are not easily defined because the group structures used in CHAT can remain ambiguous without reference back to the original source material. In XML it is only possible to represent groups of words in the transcript where they are embedded in, or isolated from, other groups. To overcome this constraint it is necessary to separate group structure from group content. Consequently, groups can be represented as isolated, embedded or overlapping. Embedded and overlapping groups are especially important in natural language processing applications.

- *Version Control for Transcripts*: A requirement necessitated by the use of partial transcription is the need to modify embellish, update improve) the transcript over time. Not only is it necessary to keep track of these modifications it is also useful too have the ability to reverse the effect of these changes. In SLA, the version control of transcripts is handled by the addition of Changes files to record updates and inserts on both the SLA Root and Descriptor files.

A fully CHAT compliant transcript could be rendered either by using an XSLT in conjunction with a helper application or by using purpose-built applications that would process the SLA XML schema directly. The latter approach is more interesting as different bespoke and third party applications could in principle contribute to the creation and management of data streams that would be registered and consolidated together to form various transcription and coding deliverables.

## 5.    Conclusions and Further Research

Spoken language is a valuable yet often underutilised resource in organisational settings. The major bottlenecks involved in utilising these resources in the study of organisations are the crucial stages of transcription and coding. These stages are time-consuming and the assumption that spoken language records need to be completely transcribed and coded reduces the likelihood that these resources will be utilised. If we first index the observational record then we can drastically reduce the amount of effort and cost of pre-processing spoken language resources by identifying only those parts of the record that need to be transcribed and coded. We applied the concept of texture from Systemic Functional Linguistics to describe and account for the effect of partial transcription on the observational record. In general, it is the text-forming resources of cohesion that are particularly sensitive to the omission of clauses and consequently this makes the resulting transcript more difficult for readers to understand as a unity. We could easily produce an unintelligible partial transcript if we did not transcribe and code enough, so knowledge of these resources provides an important theoretical underpinning to partial transcription.

One of the best transcription and coding standards is the CHAT transcription system. A critical evaluation of the TalkBank XML schema for CHAT revealed that it could not satisfy the design requirements for the spoken language architecture in organisational settings. We therefore developed a new multi-file XML schema called Spoken Language Architecture (SLA) to support CHAT while allowing transcripts to be incrementally developed, enabling us to move away from treating transcripts as monolithic units. Extensibility of the SLA schema and abstraction of the CHAT semantics address the issues of schema overloading, over-specification of the CHAT semantics and coupling between the schema and standard evident in previous approaches. Having argued the need for version control in organisational applications, the relative ease with which version control was incorporated into SLA demonstrates the appropriateness of an extensible multi-file XML schema for spoken language resources.

Future work will be involved in applying the theory of texture to account for the effects that text-forming resources will have on transcripts at various stages of completion. It is hoped that visualization tools can be developed that will enable the mutual interaction between these text-forming resources to become easier to understand and estimate. Tools that support the production of 'transcripts in pieces' are currently under development

## Acknowledgements

The research presented here was partially funded through two United Kingdom Engineering and Physical Sciences Research Council (EPSRC) grants, Semiotic Enterprise Design for IT Applications (SEDITA) jointly conducted by Staffordshire and Reading Universities, Grant Reference: GR/S04833/01, and Reducing Rework Through Decision Management (TRACKER) jointly conducted by Lancaster and Staffordshire Universities, Grant Reference: GR/R12176/01.